\documentclass[11pt]{article}
\usepackage[a4paper,margin=1in]{geometry}
\usepackage[T1]{fontenc}
\usepackage[utf8]{inputenc}
\usepackage{lmodern}
\usepackage{microtype}
\usepackage{amsmath,amssymb}
\usepackage{graphicx}
\usepackage{booktabs}
\usepackage{xcolor}
\usepackage{hyperref}
\usepackage[numbers,sort&compress]{natbib}
\usepackage{caption}
\hypersetup{colorlinks=true,linkcolor=blue!50!black,citecolor=blue!50!black,urlcolor=blue!50!black}
\graphicspath{{figures/}}
\newcommand{\ci}[2]{[#1,\,#2]}
\newcommand{\thisdoi}{10.5281/zenodo.23163677}
\newcommand{\evidencedoi}{10.5281/zenodo.23163299}

\title{Operator Packages, Proposer Strength, and\\ Construction-Family Plateaus in\\ Office-Scale Verified Search}
\author{Roberto I. Ono Filho\\ Independent researcher\\
\texttt{ono.roberto@gmail.com} \quad ORCID \href{https://orcid.org/0009-0006-8650-629X}{0009-0006-8650-629X}}
\date{Version 4, 5 October 2026 (arXiv v2). v3: 11 September 2026 (Zenodo); v2: 3 September 2026 (Zenodo); v1: 28 August 2026 (arXiv v1).\\ arXiv:2609.29636; Zenodo DOI of this version \href{https://doi.org/\thisdoi}{\thisdoi} (all versions: \href{https://doi.org/10.5281/zenodo.22280391}{10.5281/zenodo.22280391}).\\ The changes in version 4 are listed after the abstract.}

\begin{document}
\maketitle

\begin{abstract}
Verified search, in which a language model proposes programs, a hard
evaluator scores them, and selection keeps the best, has recently moved
mathematical records; controlled ablations of the proposer-side
components remain rare. We instrument a minimal FunSearch-style loop at
office scale (a 30B local model on a laptop, 120--600 verified samples
per run) with three operator packages: a schematic notebook the model
writes and carries instead of verbatim elites, a named obstacle, and
behavioural repulsion from constructions already found. On nine
construction problems from a public repository, the complete $2^3$
factorial with two replicates favours the primary contrast in a nominal
two-stage analysis: the composition closes more of the seed-to-record
gap ($+0.196$ \ci{+0.059}{+0.343}; nominal pooled $p=0.023$,
stage-combination $p\approx 0.08$; median per-problem effect $+0.045$;
robust to removing the seed floor from the metric). Repulsion raises
construction-hash diversity in eight of nine problems ($+0.31$,
one-sided $p=0.0039$; partly a manipulation check, since the operator
optimizes a relative of this quantity). The factorial finds no positive
memory-by-repulsion interaction and bounds any such interaction to about
$\pm 0.04$: the gain decomposes additively, the notebook package
carrying the largest descriptive share, and memory+repulsion is the only
arm that never collapses (0 of 18 runs), with a mean $0.025$ below the
full composition's, a difference unresolved at $n=9$. A frontier proposer under the identical
loop reaches in tens of samples what the local model does not in
hundreds, and in single scoping runs its gains arrive without the
operators (differences at most $0.001$ at equal grid). The searches
stall after closing 85--92\% of the gap toward the best published
construction on the flagship problem, and a later analysis
(Section~\ref{sec:records}, added in v3) reads that plateau as a
resolution limit: the reference value is a continuum limit and the
unaided long-run finals already carry its family's structure. In the
registered family-hint test the family named in words is adopted and
loses; handed as code, it is optimized, but our best finite-grid
encoding stays below the plateau reached unaided. We release the
harness, every candidate, and the dated pre-registrations.
\end{abstract}

{\small
\paragraph{Changes in version 4.} Corrections; no registered test or
verdict changes. (1)~Repulsion raised construction-hash diversity in
eight of nine problems (Heilbronn $-0.04$) and in 16 of 18 replicates,
not ``everywhere'' or ``in every problem and replicate'' as versions 1--3
said; $p=0.0039$ is one-sided and is the $p$ of one adverse cell, not
the minimum attainable ($1/512$). (2)~On beat-the-average the three
values compared lie at 85--92\% of the seed-to-reference gap, not
92--98\%. (3)~The abstract, Sections~\ref{sec:hint}
and~\ref{sec:discussion} and the Conclusion now agree with the note
added in version~3 (the plateau read as a resolution limit): the family
signature is absent in the three registered no-hint runs, but the same
classifier labels five of the sixteen unaided factorial finals
in-family. (4)~Grids are named in the proposer comparison: Claude's
0.3897 is at $L{=}20000$; at $L{=}5000$ its composed arm reached 0.3872
in a later run, which the 27B matches; the frontier-strength operator
differences are at most 0.001 at equal grid. (5)~The 27B was served with
vLLM on a rented H100, not locally. (6)~The note is labelled as added in
version~3, not version~2; ``the limit was not resolution'' is replaced;
three proposers, not two; nine problems, not six; 68 pages (67
problems); one typo; the provenance table gains the 27B battery and
reports the interaction as not supported; references updated.
(7)~Data availability now points to the Zenodo deposits
(Appendix~\ref{app:repro}), and a statement on the use of an AI
assistant was added.\par}

\section{Introduction}
\label{sec:intro}

FunSearch \citep{romeraparedes2024funsearch} and AlphaEvolve
\citep{novikov2025alphaevolve} demonstrated that a language model inside an
evolutionary loop with a hard evaluator can produce genuinely new
mathematical constructions. The systems that followed improved the
\emph{scaffolding}: sample-efficient archives and novelty rejection
\citep{lange2025shinka}, verbal reflections as search gradients
\citep{ye2024reevo}, stronger databases and routing. Two questions have
stayed unmeasured. First, \emph{which parts of what the proposer is asked
to think} (its memory of the lineage, its statement of the obstacle, its
relation to what was already found) change the search, and by how much?
The prompts of the published systems show elite programs verbatim and vary
everything else. Second, \emph{how much of the outcome is the proposer
itself}, holding the loop fixed? Published systems change model and
scaffold together.

This paper measures both, at a scale any laboratory can afford, with the
discipline our earlier studies \citep{ono2026interrupting} found necessary:
every battery pre-registered before running, dated amendments, exact paired
tests, and negative results reported at the same volume as positive ones.
The operators we test are not arbitrary: they are the three interventions
that survived a two-paper program on open-ended generation: a
\emph{schematic memory} (the model's own compressed recap outperformed
verbatim context on document-level integration), a \emph{standing
question}, and \emph{repulsion} from the already-produced, which in weight
space collapsed without an anchor and in behaviour space is the natural
tail-preserving form.

\paragraph{Contributions.}
(1) A pre-registered program of cognitive operators inside a fixed
verified-search loop, grown from a $5{\times}6{\times}2$ ablation to the
complete $2^3{\times}9{\times}2$ factorial: the composition reaches
$+0.196$ (nominal $p=0.023$ over the two-stage sequence; robust to the
metric's seed floor), behavioural repulsion buys construction-hash
diversity in eight of nine problems and 16 of 18 replicates
($p=0.0039$), and no single package avoids every failure mode.
(2) A revision, by the completed factorial, of this program's recurring
\emph{anchoring pattern}: the factorial finds no positive
memory$\times$repulsion interaction and bounds any such interaction to
about $\pm 0.04$ of frac; what remains of anchoring is descriptive
\emph{collapse prevention}, not superadditive gain. Repulsion without memory still breaks (three
collapses in eighteen runs; four when paired with agenda), while
memory+repulsion never does, echoing unanchored preference optimization
in weight space.
(3) A \emph{scaling inversion}: from a 120-sample tie, the composed arm
climbs to 0.3816 at 600 samples while repulsion alone stalls at 0.3514:
diversity without accumulated direction does not compound.
(4) A controlled \emph{proposer comparison}: local 30B versus a frontier
model under the identical loop and budget, on the same problem,
locating the residual gap in the proposer rather than in the loop, with a
frontier-strength scoping run in which the single-run operator
differences are at most $0.001$ at equal grid.
(5) An honest calibration of office-scale verified search: the composed
arm closes a median $0.71$ of the seed-to-reference gap (per-problem
$0.10$--$0.99$; two problems sit at plateaus every arm shares), the
tested grid and verifier-time changes did not improve the result, a
stronger proposer raised and accelerated the flagship plateau without
closing the final gap. A note added in version~3
(Section~\ref{sec:records}) reads the flagship plateau as a resolution limit of the
verifier's grid rather than a family boundary: the reference value is a
continuum limit, finite-grid members of its family cap near 0.389, and the
loop finals already carry the family's structure.

\section{Related work}
\label{sec:related}

\paragraph{Verified search systems.} FunSearch evolved single functions
with small code models and millions of samples
\citep{romeraparedes2024funsearch}; AlphaEvolve evolves whole files with
frontier ensembles and thousands \citep{novikov2025alphaevolve}, and its
repository of 67 problems with official verifiers
\citep{georgiev2025exploration} is the substrate of our experiments.
ShinkaEvolve reports record-level results with $\sim$150 samples using
novelty-based rejection and bandit model selection \citep{lange2025shinka};
ReEvo uses model-written reflections as ``verbal gradients''
\citep{ye2024reevo}. Our schematic notebook differs from ReEvo's
reflections in being a persistent, compressed lineage memory that
\emph{replaces} verbatim elites rather than commenting on pairs. A recent
wave adds persistent memory to evolutionary code search:
$\varepsilon$-MemEvo transfers compact strategy summaries across tasks
with an adaptive injection gate and reports gate ablations
\citep{liu2026memevo}; EvoMem reuses mutation knowledge across runs
\citep{volkov2026evomem}; MEMOIR shares algorithmic and failure-mode
summaries across search branches \citep{haji2026memoir}. What remains
absent, and what we contribute, is a complete factorial, within one task
suite and under a fixed local proposer, of a schematic-notebook package,
an obstacle line, and a construction-hash duplicate control. LoongFlow \citep{wan2025loongflow} composes planning, hybrid
memory and MAP-Elites niches into one system and reports large efficiency
gains; it is the closest architecture to our composed arm, and exactly the
kind of system whose components an ablation like ours aims to isolate.

\paragraph{Why diversity needs help.} Without selection pressure, LLM
mutation chains collapse into attractor regions \citep{gurkan2026mutation};
RL-tuned models lose pass@$k$ coverage relative to their base
\citep{yue2025rlvr}; and coverage under repeated sampling grows
log-linearly only when a verifier exists \citep{brown2024monkeys}. Our
behavioural-repulsion operator is novelty search
\citep{lehman2011novelty,mouret2015mapelites} transplanted into the
proposer's prompt, with the construction itself (not an embedding) as the
behaviour descriptor.

\paragraph{Automated researchers at industrial scale.} Concurrently with
this work, \citet{chen2026automated} ran LLM ``automated alignment
researchers'' that propose and train post-training fixes for ten
alignment failures. Their harness independently converges on the organs
studied here (a persistent memory file carried across fresh sessions
and a shared finding forum, whose removal costs the most headroom in
their ablation), and their telemetry shows each failure collapsing into
a single dominant method family while idea diversity stays flat or
declines as scores climb. A novelty-rejection variant raises the novelty
of accepted methods but yields mixed downstream gains, mirroring the
dissociation we certify here between diversity and value. Their setting
is complementary to ours: mitigating a well-characterized failure is a
task whose known-good region already exists in the literature, precisely
the regime in which loops of this family excel; the construction-family
plateau we document marks the boundary of that regime.

\section{The harness, the operators, the problems}
\label{sec:harness}

\paragraph{Loop.} A minimal FunSearch-style loop: $K$ islands, each holding
its best programs; per generation, per island, a prompt is built and $S$
candidates sampled; each candidate runs in a sandbox (time and memory
limits) and is scored by the problem's verifier; islands keep bounded
elites and migrate their best every third generation. The proposer is
\texttt{Qwen3-Coder-30B-A3B-Instruct} (8-bit, MLX, on a MacBook M5 Pro)
prompted through its chat template; Section~\ref{sec:records} swaps in
Claude Opus~5 through an API under the identical loop. Every candidate is
logged: code, score, error, generation, island, and a behaviour signature
(a hash of the rounded construction it outputs).

\paragraph{Operators.} \textbf{A} (baseline): the prompt shows the
statement and the island's top-3 programs with scores, FunSearch-style.
\textbf{B} (schematic notebook): each generation the model writes a
$\le$12-line notebook (what worked, what failed and why, the open
question) from the elites and the recent failures; the prompt then
shows the notebook and only the single best program. \textbf{C} (agenda):
one model-written line naming the structural obstacle of the current best,
appended to the prompt. \textbf{D} (behavioural repulsion): candidates
whose construction duplicates an island member's are rejected, and the
prompt lists already-found constructions as forbidden. \textbf{E}: B+C+D. We refer to these as operator \emph{packages}: each
bundles several changes at once (B replaces three verbatim elites with
one, adds failure summaries and an open question, and spends an extra
call; D combines a prompt-level forbidden list with a post-verification
duplicate filter), so the factorial isolates packages, not single
cognitive mechanisms; a finer within-package decomposition is registered
as future work.

\paragraph{Problems.} Nine construction problems ported from the official
repository notebooks, chosen for cheap exact verification and known
headroom. The original battery: circle packing (26 circles, maximize the
sum of radii), ring loading ($m{=}15$, exact $2^m$ verification),
beat-the-average (a pmf maximizing $P[X_1{+}X_2{+}X_3<2X_4]$), the first
autocorrelation inequality (minimize an upper bound on $C_1$),
isosceles-free subsets of the $64^2$ grid, and sum-difference III. A
pre-registered extension, run before any $n{=}9$ analysis, added three:
the max--min pairwise-distance ratio for 16 points in the plane
(minimize), the Heilbronn triangle problem for 11 points (maximize the
smallest triple area), and splitting $180!$ into 180 factors maximizing
the smallest (exact big-integer verification). Each run's budget is 120 verified
samples (10 generations $\times$ 2 islands $\times$ 6); the primary
metric \emph{frac} is the fraction of the seed-to-record gap closed,
\emph{auc} its mean over the budget (sample efficiency), \emph{div} the
share of distinct behaviours among valid candidates.

\section{The factorial}
\label{sec:factorial}

\begin{table}[t]
\centering\small
\begin{tabular}{@{}lcccccccc@{}}
\toprule
frac (pooled over 2 replicates) & A & B & C & D & BC & BD & CD & E \\
\midrule
circle packing 26 & 0.703 & 0.974 & 0.976 & 0.000 & 0.989 & 0.752 & 0.980 & \textbf{0.992} \\
ring loading 15 & 0.444 & 0.767 & \textbf{0.808} & \textbf{0.808} & 0.731 & \textbf{0.808} & \textbf{0.808} & \textbf{0.808} \\
beat-the-average & 0.181 & 0.384 & 0.519 & 0.608 & 0.661 & \textbf{0.789} & 0.540 & 0.709 \\
autocorrelation $C_1$ & 0.059 & 0.156 & 0.041 & 0.202 & 0.014 & \textbf{0.212} & 0.000 & 0.104 \\
isosceles-free $64^2$ & \textbf{0.560} & \textbf{0.560} & \textbf{0.560} & 0.399 & \textbf{0.560} & \textbf{0.560} & \textbf{0.560} & \textbf{0.560} \\
sum-difference III & \textbf{0.392} & 0.329 & 0.253 & 0.158 & 0.314 & 0.327 & 0.069 & 0.362 \\
max--min distance 16 & 0.169 & 0.992 & 0.372 & \textbf{0.993} & 0.919 & 0.793 & 0.247 & 0.708 \\
Heilbronn triangle 11 & 0.672 & 0.431 & 0.507 & 0.540 & 0.371 & 0.477 & 0.619 & \textbf{0.700} \\
$180!$ into 180 factors & \textbf{0.774} & \textbf{0.774} & \textbf{0.774} & \textbf{0.774} & \textbf{0.774} & \textbf{0.774} & \textbf{0.774} & \textbf{0.774} \\
\midrule
mean frac & 0.439 & 0.596 & 0.534 & 0.498 & 0.592 & 0.610 & 0.511 & \textbf{0.635} \\
mean construction-hash diversity & 0.41 & 0.53 & 0.52 & 0.72 & 0.59 & 0.68 & \textbf{0.80} & 0.70 \\
collapses (frac $<$ 0.05, of 18 runs) & 2 & 1 & 1 & 3 & 2 & \textbf{0} & 4 & 1 \\
\bottomrule
\end{tabular}
\caption{The complete $2^3$ factorial over nine problems, pooled over two
independent replicates (120 verified samples per run). frac = fraction of
the seed-to-record gap closed. E has the best mean and is best or
tied-best in five of nine problems; D alone is bimodal: best on wide
valid manifolds (pmfs, sequences), catastrophic where validity is fragile
(packing geometry, constrained integer sets); BD is the only arm that
never collapses, CD the one that collapses most; two problems
(isosceles-free, $180!$) are family plateaus where every arm meets the
same wall.}
\label{tab:factorial}
\end{table}

\begin{figure}[t]
\centering
\includegraphics[width=0.6\linewidth]{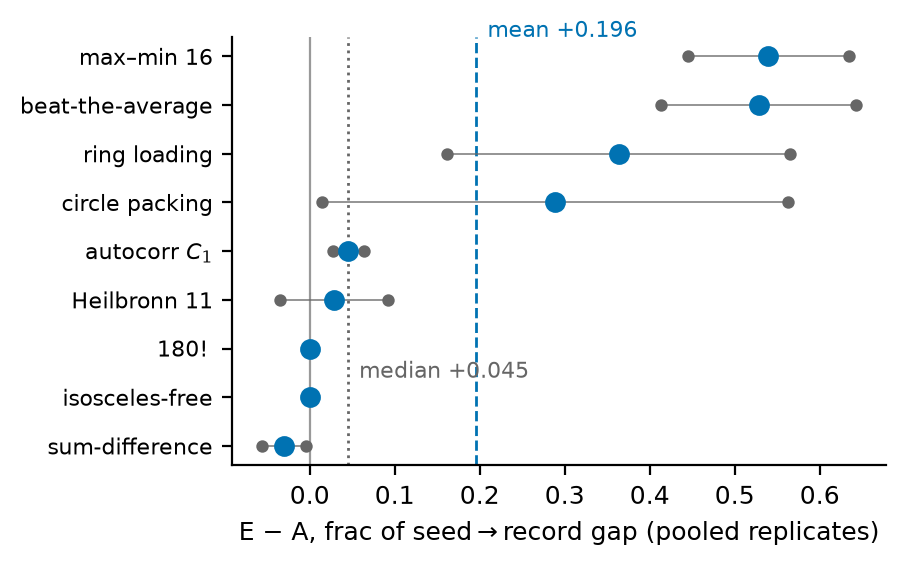}
\caption{The primary contrast, problem by problem: E $-$ A on the
fraction of the seed-to-record gap closed (pooled over two replicates).
Small grey dots are the two replicates (lines join them); the large dot
is their mean. The pooled mean is $+0.196$; the median, $+0.045$;
leave-one-problem-out means stay between $+0.153$ and $+0.224$.
Within-problem replicate spread is large on several problems (circle
packing: $+0.01$ and $+0.56$), which the problem-level test does not
see.}
\label{fig:forest}
\end{figure}

Pre-registered contrasts (exact paired sign-flip over problems, cells =
problems, pooled replicates). The original six-problem battery put the
primary contrast at the design's resolution floor ($+0.199$, $p=0.0625$;
with six cells, one adverse problem and one tie the exact test cannot go
below $4/64$); the prospectively registered nine-problem extension supports it:
\textbf{F1} (primary), E vs A on frac: $+0.196$ \ci{+0.059}{+0.343},
$p=0.023$ over the combined nine problems. We report the stages
separately, since the combination is a two-look sequence rather than an
independent confirmation: the three extension problems alone are positive
but underpowered ($+0.19$ on average, $p=0.25$, carried mostly by
max--min distance), and a Fisher combination of the stage $p$-values
(0.0625, 0.25) gives $p\approx 0.08$, so the pooled $0.023$ should be
read as nominal accumulated evidence. The result is robust to the
outcome definition: with the seed floor removed the contrast is
unchanged ($+0.196$, $p=0.023$), and excluding the two problems where
every arm meets the same plateau it rises to $+0.252$ at the same $p$
($n=7$). The effect is also heterogeneous
(Figure~\ref{fig:forest}): the median per-problem effect is $+0.045$,
leave-one-problem-out means span $+0.153$ to $+0.224$, with large gains
on four problems, little on four, and one small loss.
\textbf{F4}, D vs A on construction-hash diversity: $+0.307$
\ci{+0.149}{+0.477}, $p=0.0039$ (one-sided), positive in eight of nine
problems (Heilbronn $-0.04$) and in 16 of 18 replicates. \textbf{F6} (the anchor), E vs
D: $+0.137$ \ci{-0.047}{+0.382}, $p=0.156$: not supported as a
pairwise contrast; the factorial below resolves what it was measuring.
F2 (auc): $+0.115$, $p=0.070$: the composed arm's advantage builds late.
(The operators' measured overhead is small: one extra call per
island-generation per operator, $\sim$220 characters each, 1--3\% of
proposer output; Section~\ref{sec:limitations} gives the accounting.)

\begin{figure}[t]
\centering
\includegraphics[width=0.62\linewidth]{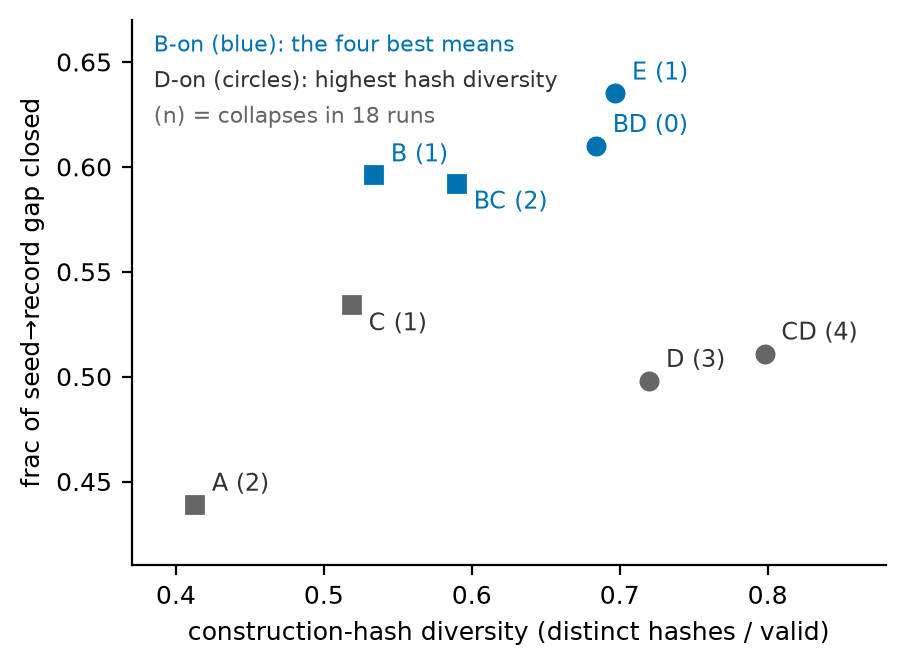}
\caption{The completed $2^3$ at a glance (cell means over nine problems,
two replicates). Memory-on cells (blue) hold the four best gap fractions;
D-on cells (circles) hold the highest construction-hash diversity; the
two axes barely interact ($\beta_{M\times R} \approx 0$). Collapse counts
in parentheses: memory+repulsion is the only arm that never collapses;
agenda+repulsion collapses most.}
\label{fig:cube}
\end{figure}

\paragraph{The completed $2^3$.}
\begin{sloppypar}
Three pairwise arms (memory+agenda, memory+repulsion, agenda+repulsion)
over the same nine problems and two replicates complete the cube
($8$ cells $\times\,9\times2$; pre-registered before running). The
effect-coded analysis (exact sign-flip over problems, $2^9$) finds no
evidence for the primary interaction: memory$\times$repulsion $\beta = +0.003$
\ci{-0.039}{+0.035}, $p=0.47$ one-sided; agenda$\times$repulsion is null
($-0.007$, $p=0.79$); and no main effect survives BH correction:
schematic memory is the largest single term ($\beta=+0.057$
\ci{+0.014}{+0.105}, raw $p=0.047$, $q=0.14$), agenda ($+0.016$) and
repulsion ($+0.011$) near zero on frac. The decomposition is exact
($E-A = 2\,(\beta_M+\beta_A+\beta_R+\beta_{MAR}) = 0.196$): the
nominal pooled composite gain splits as roughly $58\%$ notebook, $16\%$
agenda, $12\%$ repulsion and $14\%$ three-way (a descriptive
decomposition with wide per-term intervals); at $n=9$ the total is
directionally supported, the parts remain unresolved. Two
patterns in the cube are worth the trip. First, each
operator has one job (Table~\ref{tab:factorial}, Figure~\ref{fig:cube}): the four memory-on
cells hold the four best mean fracs, while every repulsion-on cell sits
at $0.68$--$0.80$ construction-hash diversity against $0.41$--$0.59$ without
it; directionally, B-on cells hold the higher means and D-on cells the
higher hash diversity, with the caveat that no main effect survives
correction.
Second, the anchor lives in the tail, not the
mean: memory+repulsion is the only arm that never collapses (0 of 18
runs, against 3 for repulsion alone and 4 for agenda+repulsion, the
cube's worst), so what memory buys repulsion is insurance against
starving the island, not superadditive mean gain. Agenda, decomposed at
last, pays for neither: near-zero on frac alone, and the worst collapse
count when paired with repulsion. Exploratory pairwise contrasts back the
parsimony reading: E vs BD $+0.025$ ($p=0.55$, median $0.000$), E vs B
$+0.039$ ($p=0.47$), BD vs B $+0.014$ ($p=0.81$), all null at $n=9$. The
collapse asymmetry is robust to the threshold: at frac $\le 0$, $<0.01$,
$<0.05$ and $<0.10$ the counts stay 3/3/3/3 for repulsion alone and
3/4/4/4 for agenda+repulsion against 0/0/0/0 for memory+repulsion, and
no run anywhere produced zero valid candidates, so these are performance
collapses, not validity collapses.
\end{sloppypar}

The heterogeneity is the finding the factorial adds to the literature:
\emph{where} the valid manifold is wide (a pmf, a sequence of pairs),
repulsion alone is the best single operator; where validity is fragile
(non-overlapping circles, exact integer constraints), it is associated
with performance collapses (both replicates of circle packing), and the composed
arm inherits most of the benefit with none of the catastrophes. No
single package avoids every failure mode; memory+repulsion was the only
configuration that avoided all of them in these runs.

\section{Scale, proposers, and the wall}
\label{sec:records}

\begin{figure}[t]
\centering
\includegraphics[width=0.86\linewidth]{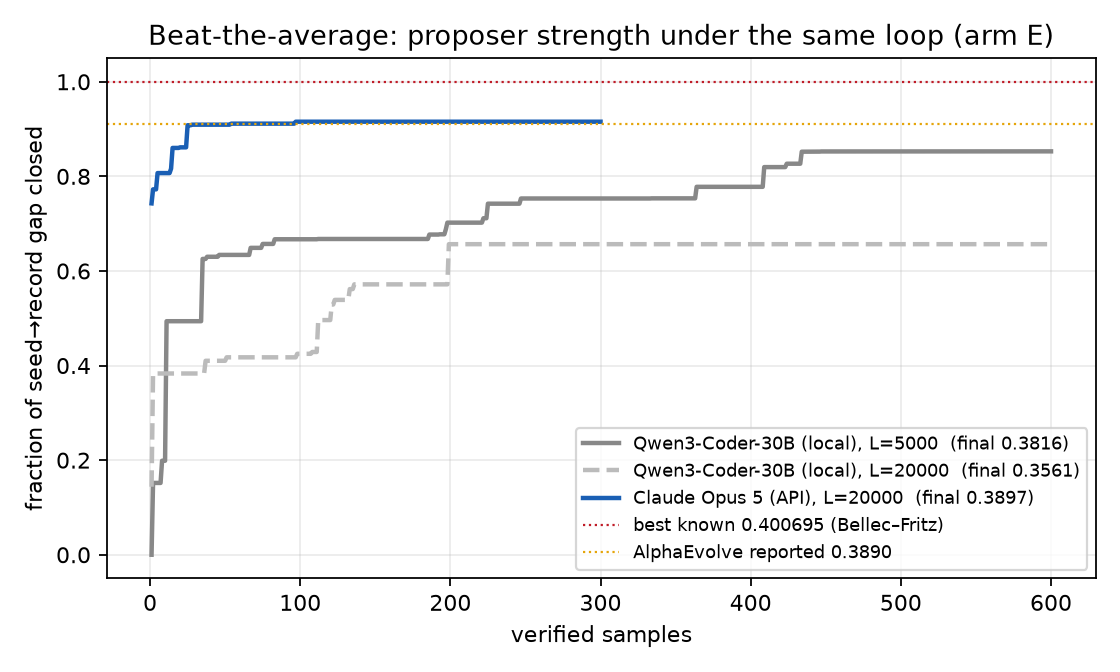}
\caption{Beat-the-average under the same loop (arm E): fraction of the
seed-to-record gap closed per verified sample. The local 30B at two grid
resolutions; Claude Opus~5 at $L{=}20000$; dotted lines mark AlphaEvolve's reported value
(0.3890) and the best known (0.400695, Bellec--Fritz).}
\label{fig:curve}
\end{figure}

We then took the problem with guaranteed headroom, beat-the-average,
where AlphaEvolve's own reported value (0.3890) sits below the best known
(0.400695), and ran a pre-registered ladder. (i) \emph{Budget}: at 600
samples the composed arm reaches 0.3816 while repulsion alone, which had
tied it at 120 samples, stalls at 0.3514: diversity without accumulated
direction does not compound. (The same $5\times$ extension on max--min
distance 16 left both local arms \emph{worse} than their own 120-sample
runs on a fresh seed: seed variance dominates budget there too.)
(ii) \emph{Grid}: refining the pmf grid from
$L{=}5000$ to $L{=}20000$ at equal budget makes the local proposer
\emph{worse} (0.3561): a finer grid did not help this proposer's search
(the note below explains why the ceiling itself rises only slowly with
the resolution). (iii)
\emph{Proposer}: Claude Opus~5 under the identical loop reaches 0.3897
(at $L{=}20000$)
within $\sim$100 samples (just above the repository's published value,
obtained with 2025-era proposers; we read this as locating the
wall, not as ranking systems across model generations) and plateaus
there for 17 generations. (iv) \emph{Verifier time}: tripling the
sandbox budget to 60 seconds, so heavier optimizers can run, moves the
plateau by $+0.0002$. (v) \emph{Proposer generation} (pre-registered
on 1 September 2026, after version~1): a current-generation open-weight
proposer, Qwen3.8-27B served with vLLM on a rented H100 (search and
verifier local) in non-reasoning mode with a 3000-token budget (the
earlier proposers had 1400), under the identical 600-sample loop at
$L{=}5000$ reaches 0.3876, between the local model (0.3816) and Claude
at $L{=}20000$ (0.3897); at $L{=}5000$ Claude's composed arm reached
0.3872 in a later run, so at equal grid the 27B matches it. Its best
construction is the same sparse-atom family, now with the exact
objective, L-BFGS weight refinement and coordinate descent on atom
positions; 60\% of its valid samples exceed the local model's whole
plateau, its generation zero shows no near-plateau (best 0.351, like
the local model and unlike Claude), so the gain is the search's, and
its validity rate is 61\% (wrong-length outputs and timeouts). The
wall survived a newer proposer by 0.013.

The constructions are legible. The local model's best is a bimodal sparse
measure of 16 atoms (half the mass on $\{0,1,2,3\}$, geometrically
spaced atoms at the top), with the strategy stated in comments. Claude's
best implements an exact sparse-atom objective with an analytic gradient
and a multiplicative-weights refinement. The frontier proposer and AlphaEvolve's published value sit within
0.0007 of each other; the local model, at $L{=}5000$, stalls about 0.008 behind them.
All three lie at 85--92\% of the seed-to-reference gap toward a
reference whose construction we read, at the time, as lying outside the
sparse-atom family every language model proposes (the note below
revises that reading). More precisely: the
candidates of all three proposers are small finitely supported measures found
numerically, while the \citet{bellec2024beat} construction is a
multiscale, recursively structured sequence of discrete measures whose
analysis shows a finitely supported measure cannot be optimal; the
family-hint conditions below name that structure. Within the settings
tested, grid and time changes did not move the result and two stronger
proposers moved it without closing the gap; these plateaus motivated the
hypothesis that what remained was the construction family, an idea. The
family-hint experiment below tests and qualifies that interpretation.

\paragraph{Note added in v3: the wall is the grid's resolution.} A later
analysis of the archived beat-the-average finals (3--4 September 2026;
appendix and scripts in Zenodo 10.5281/zenodo.22713807,
Appendix~\ref{app:repro}) changes the reading of this plateau. The
reference value 0.400695 is the continuum limit of a self-similar scale
mixture \citep[Prop.~3.3]{bellec2024beat}; the single dyadic ladder
gives 0.3804 in the continuum and 0.3814 with optimised weights.
Hand-written finite-grid members of the mixture family score 0.3850 at
$L{=}5000$, 0.3867 at $L{=}20000$ and 0.3888 at $L{=}4{,}000{,}000$,
rising slowly with the resolution. The loop finals of the 27B at
$L{=}5000$ (0.3876) and of the frontier proposer at $L{=}20000$ (0.3899,
a later run) already carry the family's structure (a dyadic ladder at
the top, a smaller copy of the ladder near zero, and split atoms that
break the ties the continuum construction exploits) and exceed our
hand-written members at the same grid. The structure is read by
inspection of the atoms: under the signature classifier registered in
Section~\ref{sec:hint}, of these finals only the local 30B's (0.3816,
600 samples) receives the ladder label, while the 27B's and the
frontier proposer's fall in the sparse-other class; so ``no unaided run
matched the signature'' (the three registered no-hint runs, by the
classifier) and ``the finals carry the structure'' (the long-run
finals, by inspection) are both true. The family was therefore not
outside what the proposers produce; what none of them can reach on a
finite grid is the continuum limit. This also explains the twist of
Section~\ref{sec:hint}: our strongest hand-written encoding, 0.374, sits
below the mixture's 0.385 at $L{=}5000$, so the family handed as code was
a weaker member of a family the loop already had. On this problem the
binding constraint is the resolution of the grid, not an idea; the other
two shared plateaus (isosceles-free, $180!$) remain candidate family
boundaries. To our knowledge, no language-model system has reported
more than about 0.390 on this problem. On other problems (circle packing, max--min distance) a
bandit study of compute allocation over evolutionary trajectories with
open models measured the same lock-in to the first family a run reaches
\citep{xing2026base}.

A second problem sharpens where such ideas live. On max--min distance 16,
the frontier proposer under the same loop emits \emph{in generation
zero}, before any search feedback, a minimax program (SLSQP on
$\min t$ s.t.\ $d^2_{ij}\le t$, $d^2_{ij}\ge 1$, analytic Jacobians)
whose value lies marginally below the repository's published one; every
valid sample across the run converges to the same value. The loop
verified this construction; it did not discover it. We treat the datum as
scope, not as a record claim: the operators structure search and protect
its tails, but they do not inject construction knowledge the proposer
lacks. The pre-registered scoping runs make the same point from the
other side, on all three registered problems. On beat-the-average, the
bare, memory-only and repulsion-only arms land within $0.001$ of one
another ($0.3871$, $0.3866$, $0.3876$; single runs, estimation only) and
within $0.001$ of the composed arm at the same grid ($0.3872$; the
$0.3897$ above is at $L{=}20000$). On max--min distance and
on the Heilbronn problem every arm converges to an identical value
($12.889230$ and $0.036530$ respectively; the Heilbronn reference in our
registry floors the reported figure, so that value reaches the frozen
reference and supports no exceedance claim). At frontier strength the
single-run operator differences are at most $0.001$ at equal grid
everywhere we measured them: the nominal gains are a local-proposer phenomenon,
and whether the collapse insurance survives at this strength remains
untested (single runs; the frontier proposer's own run-to-run noise is
unestimated).

\section{The family-hint test: verbal specification versus executable seed}
\label{sec:hint}

The registered family-hint experiment (five conditions $\times$ three
seeds, 120 samples each, memory+repulsion arm, beat-the-average) asks
what it takes to move the search into the \citet{bellec2024beat}
family: (a) no hint; (b) a generic hint; (c) a structural hint naming
the family without parameters; (d) a crude in-family seed program
(initial score 0.2734); (e) a \emph{strong in-family seed}, our best
tested finite-grid instantiation of the family (0.3742; on this grid
with strict inequality, notably \emph{below} the 0.3816 plateau of the
unaided 600-sample run, a fact measured and registered before launch;
the condition supplies an author-implemented finite approximation, not
the theoretical multiscale construction itself).

\begin{table}[t]
\centering\footnotesize
\setlength{\tabcolsep}{5pt}
\begin{tabular}{@{}lccccc@{}}
\toprule
condition & seed 11 & seed 12 & seed 13 & mean & finals in-family \\
\midrule
no hint & 0.3496 & 0.3607 & 0.3463 & 0.3522 & 0/3 \\
generic hint & 0.3512 & 0.3381 & 0.3086 & 0.3326 & 0/3 \\
structural hint & 0.3284 & 0.3383 & 0.3140 & 0.3269 & 3/3 \\
crude in-family seed & 0.3554 & 0.3679 & 0.3520 & 0.3584 & 2/3 \\
strong in-family seed & 0.3762 & 0.3739 & 0.3743 & 0.3748 & 3/3 \\
\bottomrule
\end{tabular}
\caption{Family-hint test: final best per seed. In-family membership by
the preregistered family-signature classifier (at most 50 support atoms,
mass $\ge 0.3$ on $\{0..3\}$, atoms in $\ge 3$ $\log_{10}$
distance-from-top bands; the structural hint instructs exactly this
structure, so the classifier is partly a manipulation check of
instruction-following). With three seeds per condition every contrast is
descriptive (the smallest one-sided sign-flip $p$ at $n{=}3$ is 0.125).}
\label{tab:hint}
\end{table}

The answer came with a twist (Table~\ref{tab:hint}). The structural
hint \emph{worked as an instruction and lost as a search}: its final
constructions carry the family signature, so the model obeyed, entered a
family that is inferior at this grid, and paid about $0.025$ for it.
Handed the family as code, the loop optimizes it ($+0.085$ from the
crude seed, above the no-hint endpoint in three of three seeds) and
stays loyal to it (no strong-seed run deserted to the sparse
constructions of the unaided runs). No unaided candidate in the three
registered no-hint runs matched the family signature. Two caveats bound the reading: the
historical $0.3816$ comes from a different battery under strong seed
variance, so the strong seed's $0.375$ exceeds the \emph{contemporary}
no-hint $0.352$ descriptively while trailing the historical best; and
the seed conditions enter with different initial scores, so final values
mix family advantage with starting-point advantage (matched-score sparse
seeds and paraphrased or deliberately wrong hints are the registered
next controls). Within those bounds the registered hypothesis was not
supported: handing the idea, in words or in this finite encoding, did
not cross the last mile under this budget; our tested finite-grid
encoding of the idea was not sufficient to close the gap, which leaves
open whether the theoretical family, a richer encoding, or a larger
budget would. What the experiment does establish is a division of
labour: under these runs the loop transported and optimized the idea it
was handed and did not produce its signature unaided, and instructing a
construction is not inducing a competitive implementation of it, the
third appearance of that pattern in this program (the continuity
connective of the companion study; the frontier scoping runs; now the
structural hint).

\section{Discussion}
\label{sec:discussion}

\paragraph{What office scale buys.} With 120--600 verified samples, a
local model and one laptop: a median $0.71$ of the seed-to-reference
gap closed by the composed arm across nine problems (per-problem
$0.10$--$0.99$), legible programs, and, with a frontier proposer
at $\sim$US\$50 of API, parity with the values the large systems report on
this problem. The published record-movers used frontier ensembles with
thousands of samples \citep{novikov2025alphaevolve}, millions with small
models \citep{romeraparedes2024funsearch}, or open-source loops with
strong API models \citep{lange2025shinka}; the band we measure (local
proposer, hundreds of samples, controlled operators) was unoccupied,
and it is the band most laboratories can afford.

\paragraph{The operator ledger.} E had the highest mean and one
collapse; memory+repulsion was the only zero-collapse arm, at a mean
within $0.025$ of E's and without the agenda's extra call, so the
parsimonious operational recipe these data support is memory+repulsion,
with the full composition as the highest-mean variant. The mechanism is
visible in the notebooks: the notebook package carries lineage
information across generations and the duplicate control spends the
budget on genuinely
different constructions. Alone, each fails somewhere: memory and agenda
are inert on problems the model already saturates; repulsion is
associated with performance collapses on fragile manifolds. This is a recurring collapse-prevention pattern,
consistent with the anchoring hypothesis, that this program has now
observed across narrative generation, weight-space preference
optimization, and prompt-space search.

\paragraph{The proposer comparison and the plateau.} On this one problem,
the frontier proposer is far more efficient in verifier evaluations
($+0.008$ absolute over the local model's 600-sample best, at
$L{=}20000$ against $L{=}5000$, and $+0.006$ at equal grid in a later
run; reached with roughly $20\times$ fewer verified samples), and
everything we tried after it moved little (grid, time, more samples:
$+0.0002$). This is a sample-efficiency comparison on one problem under
one harness, not yet a proposer-strength curve; replications on further
problems were planned but not run, and reasoning effort, ensembles, alternative
representations and thousand-sample budgets all remain untested
explanations. The shared plateau suggests, consistently with our
consolidation results \citep{ono2026mass}, that search over a fixed prior
rapidly collects what the prior's construction families contain and then
stalls at the best family member, a hypothesis to test rather than
assert. Section~\ref{sec:hint} put that hypothesis to a registered test; the
short answer is that a verbal specification of the reference family was
followed and lost, while an executable in-family seed was optimized, and
no unaided candidate in the three registered no-hint runs produced the
family signature (applied later to all archived finals, the same
classifier labels five of the sixteen unaided 120-sample factorial
finals in-family). (The note in
Section~\ref{sec:records} revises this for the flagship problem: the long-run
finals do carry the family's structure, and the plateau is the grid's
resolution limit.) This is the question
\citet{georgiev2025exploration} report answering with an expert in the
loop, from the other side.

\paragraph{Design rules.} For verified search at small scale: pair
memory with repulsion (the zero-collapse arm; the full composition adds
$+0.025$ of mean at the cost of the agenda's extra calls, a difference
our data cannot distinguish from zero), and never run repulsion alone on
fragile manifolds; keep the
behaviour descriptor construction-derived (a rounded construction hash,
not an embedding; no symmetry canonicalization); spend on
the proposer before spending on samples; treat plateaus as candidate family
boundaries and test them (grid, time, proposer and hints in
pre-registered steps) rather than declaring them.

\section{Experimental provenance}
\label{sec:provenance}

Every battery was registered before running; the combined analyses are
sequential, and the table makes the order of looks explicit.

\begin{table}[h]
\centering\small
\begin{tabular}{@{}lllll@{}}
\toprule
battery & registered & data already seen & primary & status \\
\midrule
F ($5{\times}6{\times}2$) & 2026-08-22 & none & E$>$A frac & $p=0.0625$ (floor) \\
F-ext ladder & 2026-08-23 & F & budget/grid/proposer & descriptive \\
F-ext2 ($n{=}9$) & 2026-08-24 & F, ladder & E$>$A frac & $p=0.023$ combined \\
$2^3$ completion & 2026-08-24 & F, ext2 partial & M$\times$R $>0$ & not supported ($p=0.47$) \\
Frontier scoping & 2026-08-26 & all above & estimation only & operators $\approx$ null \\
Family hint & 2026-08-27 & all above & optimize-once-given & yes as code, no as words \\
27B proposer (v) & 2026-09-01 & all above & wall holds vs breaks & holds (0.3876) \\
\bottomrule
\end{tabular}
\caption{Order of registrations and looks. The nine-problem primary
analysis is a two-look sequence (discovery + prospective extension), not
an independent confirmation.}
\label{tab:provenance}
\end{table}

\section{Limitations}
\label{sec:limitations}

Nine problems is still a small universe of construction families; the
full $2^3$ isolates interactions, but at $n{=}9$ no per-term effect
survives correction, so the operator decomposition (memory largest) is
directional, not statistically resolved, and the collapse asymmetry is
descriptive;
budgets are equalized in proposals and verifier evaluations (120 of each
per run; duplicate rejection in the repulsion arms happens after
verification, since the signature requires the construction, and discards
20--30 verified candidates per run from the islands); the operators'
extra calls are measured at one per island-generation per operator
($\sim$220 characters per call, an overhead of 1--3\% of proposer
output characters; input-token accounting is incomplete, so budgets are
matched in proposals and verifications, not demonstrably in tokens), and
the arms differ in candidates surviving dedup into the islands (per run:
A 108, B 95, C 93, D 72, BC 94, BD 65, CD 73, E 63), so sample
efficiency here means verifier-evaluation efficiency; the diversity
metric is related to the repulsion operator's own filter; the behaviour
signature is a hash of the returned construction after rounding, with no
canonicalization for symmetries (permutation, rotation, reflection), so
div can count symmetric variants as distinct, its rounding sensitivity is
untested, and the D filter and the div metric share this signature (a
retrospective AST-structure metric, invariant to renaming and
independent of that signature, is directionally consistent on the runs
with logged code: D-on arms 0.67--0.87 distinct structures per valid
candidate against 0.38--0.39 for A and B on the matched problems; paired
D$-$A $+0.27$ at the $n{=}3$ resolution floor); the nine problems (six
chosen before any runs, three added by the pre-registered extension under
the same criteria) satisfy criteria fixed before any runs (cheap exact verifier,
short-function candidates, known headroom; the universe of 68 candidate
pages (67 problems)
and per-problem ratings are in the released reconnaissance table) but were
not sampled randomly, so the exact tests quantify sign consistency
within this selected suite and do not license random-sample
generalization to the repository or to construction problems generally;
because the problems and their literature are public, proposer
performance does not distinguish retrieval of pretrained construction
knowledge from de novo discovery (the generation-zero minimax program is
the extreme case); \emph{frac} is floored at the seed by construction,
signs are harmonized for minimization, and reference values are frozen at
the repository commit recorded in the release; the scaling ladder is one
domain; the API proposer is not bitwise reproducible and its reasoning
effort was fixed low; comparisons against repository values compare a
2026 proposer with values obtained by 2025-era systems, so they locate
our wall and never rank systems: no system-level head-to-head with
ShinkaEvolve or AlphaEvolve is attempted, and the operators should be
read as structuring search, not as a substitute for construction
knowledge the proposer lacks.

\section{Conclusion}
\label{sec:conclusion}

Inside a fixed verified-search loop at office scale, the operator
packages our generation studies isolated (a schematic notebook, a named
obstacle, behavioural duplicate control) compose into the arm with the
highest mean, E, at one collapse in eighteen runs; memory+repulsion was
the only zero-collapse configuration, within $0.025$ of E's mean, and is
the parsimonious recipe these data support. Repulsion alone enforces
behavioural non-duplication by construction and was the best single
package exactly where the valid manifold is wide; and on the one problem
where we compared them, a frontier proposer reached a higher value with
roughly $20\times$ fewer verifier evaluations. On beat-the-average all
three proposers stalled below a multiscale reference whose value is a continuum
limit; the note added in version~3 reads that plateau as the resolution limit of
the grid, with the reference's family already in the loop's finals, and
the other plateaus as candidate family boundaries not yet established.
Whether crossing such a boundary requires what we have informally called an idea
was put to a registered test on the flagship problem: the idea, handed
in words, was followed and lost; handed as code, it was optimized, but
our best tested finite-grid instantiation remained below the plateau the
models reach unaided. In the registered test the loop moved the idea it
was handed, and its three no-hint runs produced no candidate matching
the idea's signature; our finite encoding of the idea was not sufficient
to close the gap here. Whether the last mile of these searches is an idea, the
question this program has carried informally, remains open, now with its
first measured answer.

\paragraph{Acknowledgements and use of AI.} The experiments, analyses
and text of this paper were produced by the author working with an AI
assistant (Claude, Anthropic) inside the Claude Code environment: the
assistant wrote and ran the code, operated the runs under the author's
direction and drafted the manuscript. Per the commit trailers, the
assistant models were mainly Claude Fable~5 (August 2026) and Claude
Fable~5.1 (September 2026); the corrections of version~4 were made with
Claude Opus~5.5. Every design decision, result and claim was reviewed
by the author, who is responsible for the content. Before version~1 a
colleague read the drafts in five rounds (24--28 August 2026), and the
comments were applied before submission; the reading adjustments of
version~3 came from a cross-reading of the series by another AI-assistant
session, checked against the sources. Neither is peer review,
and the paper has not been peer reviewed. Claude Opus~5, the frontier
proposer under test, belongs to the experiments, not to the tooling.

\appendix
\section{Reproducibility and data availability}
\label{app:repro}

\begin{sloppypar}
Code: \url{https://github.com/RobertoOno/interrupting-the-loop}, the
repository of the program, with the search
harness (\path{scripts/frontier_search.py}), the problem verifiers
(\path{src/creative_machine/domains/}), the batteries
(\path{run_f*.sh}, \path{run_hint.sh}), the analyses
(\path{frontier_analysis.py}, \path{analysis_23.py}), the
generated appendices (\path{docs/APPENDIX_F*.md},
\path{APPENDIX_HINT.md}) and the dated laboratory notebook
(\path{docs/PLANO.md}, in Portuguese) with the pre-registrations and
amendments of the batteries through the family-hint test. Run data and
the later notebook entries are in three Zenodo deposits.
(i)~The release \texttt{run-data-v1} of the repository, deposited with
\texttt{run-data-v2} as
\href{https://doi.org/10.5281/zenodo.22714138}{10.5281/zenodo.22714138}:
every logged candidate (\texttt{history.jsonl}, \path{best.py}) of the
factorial with its replicate and extension batteries, of the frontier
scoping runs and of the family-hint test.
(ii)~The evidence deposit of the later programs of this series
(programs 5--7),
\href{https://doi.org/10.5281/zenodo.22713807}{10.5281/zenodo.22713807}:
the ladder runs, rungs (i)--(iv) in \path{runs/frontier/record} and
the proposer-generation rung (v) in \path{runs/frontier/record27}
(with \path{run_record27.sh}); the scoping runs again
(\path{runs/frontier/FP}); and the analysis behind the note in
Section~\ref{sec:records} (\path{docs/APPENDIX_P2_FAMILY.md},
\path{family_beatavg.py}, \path{bf_ladder_limit.py},
\path{bf_mixture_grid.py}) with its dated notebook entry.
(iii)~The evidence deposit of this paper,
\href{https://doi.org/\evidencedoi}{\evidencedoi}: the frontier
proposer's later composed-arm run at $L{=}5000$
(\path{runs/frontier/claude/beatavg_E300}) and the notebook entries
made after version~1 that the other deposits lack (the pre-registration,
amendments and single read of rung (v), and the entries behind the
equal-grid comparison). Together the three deposits hold every
candidate of every reported run. Proposing used
\texttt{Qwen3-Coder-30B-A3B-Instruct} (8-bit, MLX) locally,
\texttt{Qwen3.8-27B} served with vLLM on a rented H100 for rung (v),
and \texttt{anthropic.claude-opus-5} on Amazon Bedrock for the frontier
comparison.
\end{sloppypar}

\bibliographystyle{plainnat}
\bibliography{references}

\end{document}